\documentclass{article}
\usepackage{spconf,amsmath,graphicx}
\usepackage{multicol, multirow}
\usepackage{graphics, graphicx} 
\usepackage{epsfig} 
\usepackage{mathptmx} 
\usepackage{times} 
\usepackage{booktabs}
\usepackage{caption}
\usepackage{color}

\def\ours{MEDeA} 

\title{MEDeA: Multi-view Efficient Depth Adjustment}
\name{Mikhail Artemyev, Anna Vorontsova, Anna Sokolova, and Alexander Limonov
}
\address{Samsung Research
        }
\begin{document}
%
\maketitle
\begin{abstract}
The majority of modern single-view depth estimation methods predict relative depth and thus cannot be directly applied in many real-world scenarios, despite impressive performance in the benchmarks. Moreover, single-view approaches cannot guarantee consistency across a sequence of frames. Consistency is typically addressed with test-time optimization of discrepancy across views; however, it takes hours to process a single scene. In this paper, we present \ours, an efficient multi-view test-time depth adjustment method, that is an order of magnitude faster than existing test-time approaches. Given RGB frames with camera parameters, \ours{} predicts initial depth maps, adjusts them by optimizing local scaling coefficients, and outputs temporally-consistent depth maps. Contrary to test-time methods requiring normals, optical flow, or semantics estimation, \ours{} produces high-quality predictions with a depth estimation network solely. Our method sets a new state-of-the-art on TUM RGB-D, 7Scenes, and ScanNet benchmarks and successfully handles smartphone-captured data from ARKitScenes dataset.
\end{abstract}


\section{INTRODUCTION}
\label{sec:intro}

Depth estimation is a core technology for image and video processing, serving many downstream tasks, such as 3D scene reconstruction, video stabilization, applying a bokeh effect, etc.
Compared to other geometric representations, such as voxels, point clouds, implicit neural representations, or truncated signed distance functions (TSDF), representation in a form of 2D depth maps is memory-efficient and suitable for real-time processing.

\begin{figure}[ht!]
\centerline{\includegraphics[width=.9\columnwidth]{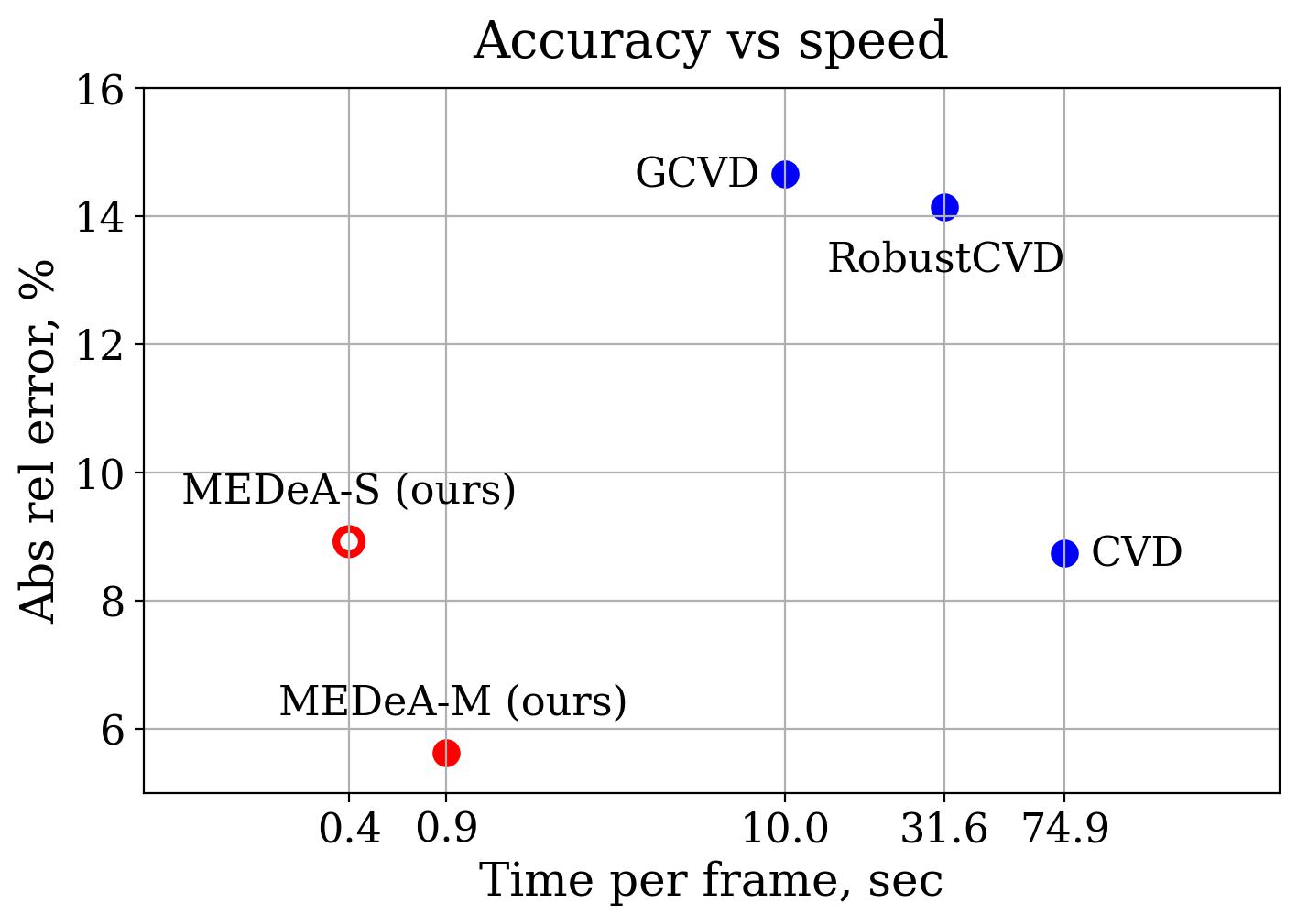}}
\caption{Comparison of depth estimation errors and runtime on the TUM-RGBD dataset~\cite{dehghan2021arkitscenes}. The proposed \ours{} surpasses existing test-time optimization methods CVD~\cite{luo2020cvd}, RobustCVD~\cite{kopf2021rcvd}, and GCVD~\cite{lee2022gcvd}) in both accuracy and speed. Our flagship \ours{}-M model outperforms competitors by a huge margin, and even our fast \ours{}-S delivers higher quality than existing approaches with a 25x speed up. Seconds per frame are in logarithmic scale for visibility.}
\label{fig:teaser}
\end{figure}

Existing single-view depth estimation models predict depth either up-to-shift-and-scale, up-to-scale, or in absolute values (metric). 


Both up-to-shift-and-scale~\cite{ranftl2020midas} and up-to-scale~\cite{romanov2021gp} methods are inconsistent by design: the depth scale may fluctuate significantly throughout a video sequence, so predicted depth maps do not align in 3D space.
Metric depth estimation methods provide scale-aligned outputs across a set of frames, yet processing frames individually inevitably leads to inconsistent predictions. Accordingly, single-view depth predictions require additional alignment if aiming at video depth estimation. 

Several video depth estimation methods actually follow this paradigm, performing test-time optimization of depth priors obtained with a single-view model~\cite{luo2020cvd, kopf2021rcvd, lee2022gcvd}. Currently, inference speed is the main limitation of test-time optimization methods, making real-time performance unattainable. We are the first to propose the test-time depth optimization method that is on par with feed-forward approaches in speed.

Another way to use information from several frames is a multi-view stereo (MVS) paradigm. Given camera parameters, MVS approaches~\cite{sayed2022simplerecon, huang2018deepmvs,yao2018mvsnet,long2021multiview, duzceker2021deepvideomvs} estimate metric depth. However, since they do not minimize between-frame discrepancy directly, thus cannot guarantee temporal consistency.

Our main contributions are as follows:
\begin{itemize}

\item We introduce \ours{} for \textbf{fast test-time video depth estimation}, which is on par with feed-forward models in speed, and is an order of magnitude faster than competing test-time optimization approaches; 

\item We show that consistent video depth estimation can be addressed with a \textbf{minimal pipeline without any auxiliary modules} for optical flow estimation, surface normal estimation, or segmentation;

\item With our \textbf{novel depth scale propagation strategy}, which enforces the multi-frame coherence and speeds up the convergence, \ours{} sets a new state-of-the-art in video depth estimation. It outperforms existing single-view, MVS, and test-time depth estimation methods in the TUM RGB-D, 7Scenes, and ScanNet benchmarks, while the experiments on the ARKitScenes show that \ours{} can handle smartphone data available in mobile applications.

\end{itemize}

\section{RELATED WORK}
\label{sec:related}

We overview existing depth estimation methods that are applicable for videos: single-view, multi-view, and test-time optimization approaches.

\subsection{Single-view Depth Estimation}

To handle a variety of real scenes, a single-view depth estimation should generalize over diverse domains. MiDaS~\cite{ranftl2020midas} is a seminal work on general-purpose depth estimation, that proposes a scheme of training on a mixture of datasets to generalize over real data, however, it predicts depth up to an unknown shift and scale. Romanov et al.~\cite{romanov2021gp} improves MiDaS training scheme to achieve up-to-scale depth estimation with a lightweight model. Recent ZoeDepth~\cite{bhat2023zoedepth} fine-tunes MiDaS for metric depth estimation, but still experience generalization issues, similar to other approaches that predict absolute depth. Besides, ZoeDepth relies on a large backbone, resulting in a slow inference comparable to some MVS methods.

\subsection{Multi-view Depth Estimation}

MVS methods improve over single-view approaches by aggregating information from several frames. Most learning-based MVS methods apply plane-sweeping to generate a cost volume. MVSNet~\cite{yao2018mvsnet} and DPSNet~\cite{im2019dpsnet} build 4D feature volumes and process them with computationally-demanding 3D convolutions, while MVDepthNet~\cite{wang2018mvdepthnet} directly generates 3D volumes by calculating cost on images by 2D convolutions. DELTAS~\cite{sinha2020deltas} learns to detect and triangulate keypoints, and converts the sparse set of 3D points into dense depth maps.

Some learning-based MVS approaches also exploit spatial similarity of consequent frames: e.g., GP-MVS~\cite{hou2019gpmvs} extends MVDepthNet~\cite{wang2018mvdepthnet} with Gaussian Process conditioned on a similarity between camera poses, and DeepVideoMVS~\cite{duzceker2021deepvideomvs} training a spatio-temporal Conv-LSTM network for early-stage cost volume fusion. SimpleRecon~\cite{sayed2022simplerecon} processes a plane-sweep feature volume without costly 3D convolutions, achieving high efficiency. 

When estimating depth for a target frame, MVS methods do not take other depth predictions into account. This results in a low between-frame correlation of depth errors, so there is room for consistency improvement.

\subsection{Test-time Depth Optimization}

CVD~\cite{luo2020cvd} is a pioneer approach improving between-frame consistency through test-time optimization. CVD leverages COLMAP camera poses, and fine-tunes a pre-trained depth estimation model using reprojection-based losses. RobustCVD~\cite{kopf2021rcvd} freezes a pre-trained depth estimation model and jointly estimates poses and depth scale maps. GCVD~\cite{lee2022gcvd} speeds up the scale maps estimation, and ensures global consistency for long videos by integrating a keyframe-based pose graph into learning. IronDepth~\cite{Bae2022IronDepth} exploits a pre-trained normal estimation network and uses predicted surface normals to guide the recurrent refinement of depth maps.

All these test-time optimization methods apply auxiliary models for optical flow estimation and dynamic objects segmentation~\cite{luo2020cvd, kopf2021rcvd, lee2022gcvd}, or surface normals estimation~\cite{Bae2022IronDepth}. In contrast, we only use a pre-trained single-view depth estimation network. Unlike CVD~\cite{luo2020cvd} and GCVD~\cite{lee2022gcvd}, we keep it frozen to avoid costly gradient backpropagation through the model. Furthermore, we propose using the depth estimation network as a feature extractor, and additionally guide the optimization via a feature-metric loss inspired by~\cite{shu2020featdepth}.

RobustCVD~\cite{kopf2021rcvd} and GCVD~\cite{lee2022gcvd} use MiDaS~\cite{ranftl2020midas} to obtain initial depth. Yet, they adjust only the scale -- while MiDaS actually predicts inverse depth up-to-shift-and-scale, so an inverse depth shift in predictions remains uncompensated. Surprisingly, this issue has been neither empirically investigated nor discussed. In \ours, we use a lightweight up-to-scale model by Romanov et al.~\cite{romanov2021gp} for initial depth estimation, which requires 12.9x less FLOPs than MiDaS. 




\section{PROPOSED METHOD}
\label{sec:method}

\begin{figure*}[h!]
\centerline{\includegraphics[width=0.825\textwidth]{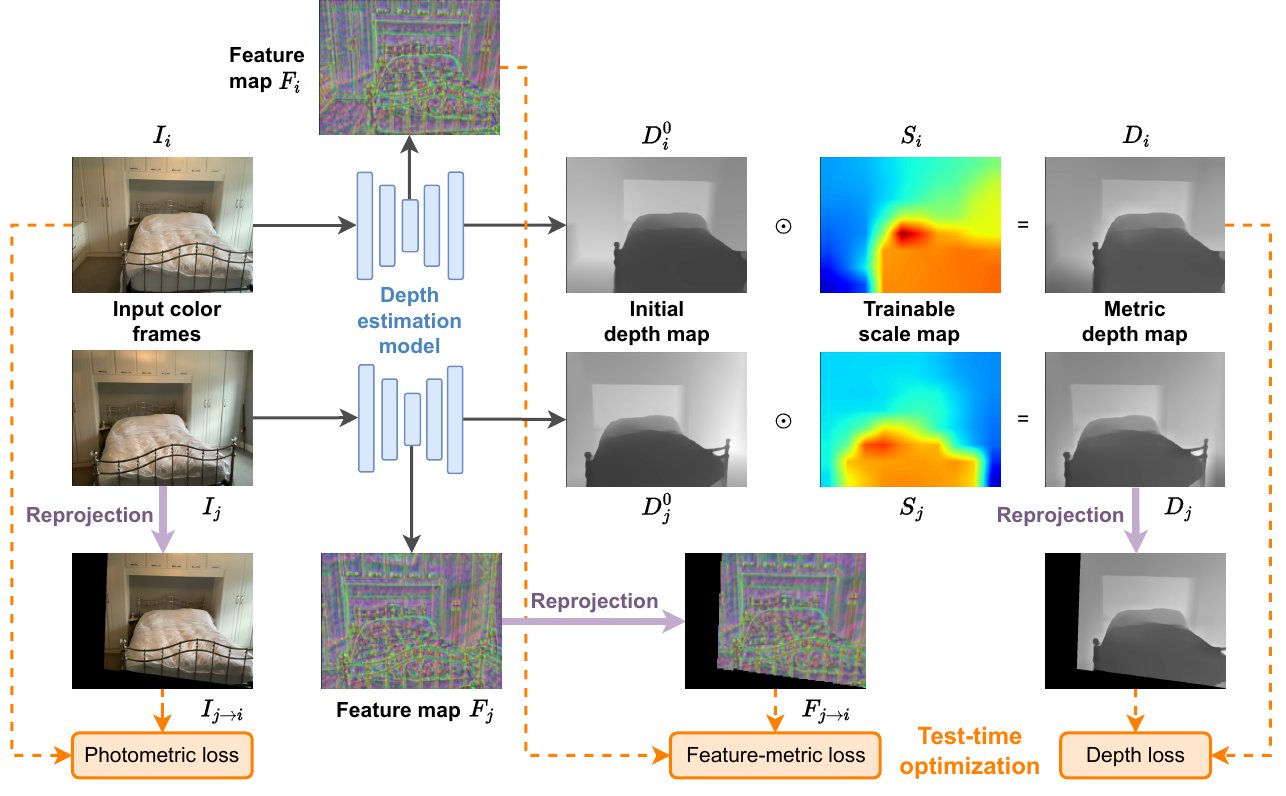}}
\caption{Overview of \ours{}. \ours{} relies on a depth estimation model that outputs either metric or up-to-scale depth maps $D^0$. 
In our depth deformation model, a depth map $D$ is estimated as an initial depth map $D^0$ multiplied by a depth scale map $S$: $D = D^0 \odot S$. At each iteration, a pair of RGB-D frames with indices $(i, j)$ is selected, and $S_i,\ S_j$ are adjusted. Using the current estimate of $D_i$, \ours{} reprojects an image $I_j$, a depth map $D_j$, and a feature map $F_j$ onto the $i$-th viewpoint, and penalizes the divergence of the reprojected $\{ I_{j\rightarrow i}, D_{j\rightarrow i}, F_{j\rightarrow i}\}$ and the original $\{ I_i, D_i, F_i \}$ values. Depth scale maps $S_i,\ S_j$ are optimized via backpropagation. As a result, \ours{} provides consistent depth maps $D_i,\ D_j$.}
\label{fig:scheme}
\end{figure*}

For the $i$-th frame, we use a pre-trained and frozen depth estimation network to estimate a depth map $D^0_i$ and extract image features $F_i$ (so that feature extraction does not require any additional computation). Then, we adjust $D^0_i$, resulting in temporally-consistent depth maps $D_i$ (Fig.~\ref{fig:scheme}). 

\subsection{Depth Estimation Network}

We leverage the efficient single-view backbone EfficientNet-b5-LRN from Romanov et al.~\cite{romanov2021gp} in \ours-S, and the accurate multi-view SimpleRecon~\cite{sayed2022simplerecon} in \ours-M. The experiments show that \ours{} is compatible with such distinct models, and robustly improves their performance in the standard tests. Since \ours{} does not impose any specific limitations, we assume that any single- or multi-view method can be incorporated into our pipeline. 

\subsection{Depth Deformation Model}

For a frame $F_i$, a \textit{depth scale map} $S_i$ is estimated in test-time optimization, and the final depth prediction is calculated as:
\begin{equation}
D_i = D^0_i \cdot S_i. 
\end{equation}
Same as RobustCVD~\cite{kopf2021rcvd} and GCVD~\cite{lee2022gcvd}, we use a parametric depth deformation model to adjust depth predictions; we keep the same simple spatially-varying depth scale model to control the computational complexity. In such a model, $S_i$ is a bilinear spline:
\begin{equation}
S_i = B(\exp(l_i), H, W), 
\end{equation}
where $l_i$ is a trainable tensor of size of $h\times w$, and $B(*, H, W)$ denotes the bilinear interpolation into the frame resolution $H\times W$. Hence, local scale coefficients are defined on a regular grid: for each pixel in a grid cell, the four values in the cell vertices are bilinearly interpolated to get a scaling factor.

\subsection{Test-time Optimization}

Our optimization relies on a reprojection induced by the depth for the $i$-th frame. Specifically, we reproject an image $I_j$, image features $F_j$, and a scale-adjusted depth map $D_j$ w.r.t the pose of $i$-th frame, using camera poses and intrinsic parameters, to get a pseudo image $I_{j\rightarrow i}$, pseudo feature map $F_{j\rightarrow i}$, and a pseudo depth map $D_{j\rightarrow i}$, respectively. Ideally, the pseudo color, depth map, and feature map should coincide with the original ones, so we penalize their divergence. 

The optimization runs in two stages: at the Stage I, depth scale maps are optimized for keyframes only. Then, the depth scale is propagated from keyframes to other frames. 
At the Stage II, all frames are involved in optimization.

\subsection{Sampling Frame Pairs}

Following CVD~\cite{luo2020cvd}, we use a hierarchical sampling scheme. Let us denote \begin{equation}
    \text{Pairs}(a, b) = \left\{(i, j)\left| \vert i - j\vert = 2^l,\ i \text{ mod } 2^l = 0, a\leq l \leq b\right.\right\} 
\end{equation}
Unlike CVD, we sample two sets of frame pairs: $P_I = \text{Pairs}(3, 6)$ (keyframes, $K=\bigcup P_{I}$), and $P_{II} = \text{Pairs}(0, 2)$. 

During the Stage I, we select keyframe pairs $(i, j)$ uniformly from $P_I$. During the Stage II, we update depth scale maps $l_i$ for $i\notin K$. We freeze parameters $\{l_i|i \in K\}$ and optimize w.r.t. frame pairs from $P_{II}$ solely.

\subsection{Depth Scale Propagation}

As a result of the Stage I, we obtain consistent depth maps for keyframes $K$. They are a valuable source of depth scale information, which can be propagated to non-keyframes. Therefore, the optimization at the Stage II can start from a reasonable scale approximation. 

For a target non-keyframe with an index $i\notin K$, we select the temporally-nearest keyframe with an index $j \in K$, and analytically derive $l_i$ that minimizes $\mathcal{L}_\text{depth}(i, j)$ (see Subsec. \ref{ss:losses}). By dividing the reprojected depth $D_{j\to i}$ by an initial depth map $D^0_i$, we obtain a depth scale map $\hat S_i=D_{j\to i}/D^0_i$. Finally, we calculate the initial parameters of the depth deformation model for the $i$-th frame as: \begin{equation}
    l_i:=\log\left(\text{Pool}(\hat S_i, h, w)\right),
\label{eq:scale-propagation}
\end{equation} 
where $\text{Pool}(*, h, w)$ denotes pooling into size $(h, w)$. We use median pooling to ensure the robustness of scale estimation.

\subsection{Losses}
\label{ss:losses}
First, the \textit{photometric loss} $\mathcal{L}_\text{photo}$ forces a pseudo color $I_{j\rightarrow i}(p)$ of a pixel $p$ to be consistent with a color $I_i(p)$: \begin{equation}
\mathcal{L}_\text{photo}(i, j) = \frac{1}{3\vert I_{j\to i}\vert}\sum_{p\in I_{j\to i}}\sum_{c=1}^3|I_i(p, c) - I_{j\rightarrow i}(p, c)|,
\end{equation}
where $\vert I_{j \to i} \vert$ is a number of valid pixels in a pseudo image $I_{j \rightarrow i}$.
Following GCVD~\cite{lee2022gcvd}, we use a \textit{depth loss} $\mathcal{L}_\text{depth}$ bringing a pseudo depth $D_{j\rightarrow i}(p)$ of a pixel $p$ closer to $D_i(p)$: \begin{equation}
\mathcal{L}_\text{depth}(i, j) = \frac{1}{\vert I_{j\to i}\vert}\sum_{p\in I_{j\to i}}|D_i(p) - D_{j\rightarrow i}(p))|
\end{equation}

The \textit{feature-metric} $\mathcal{L}_\text{feat}$ ensures that $F_{j\rightarrow i}$ and $F_i$ are alike. For \ours-S, we formulate the feature distance as: \begin{equation}
    \mathcal{L}_\text{feat}(i, j) = \frac{1}{C\vert I_{j\to i}\vert}\sum_{p\in I_{j\to i}}\sum_{c=1}^C \dfrac{|F_i(p, c) - F_{j\rightarrow i}(p, c)|}{|F_i(p, c) + F_{j\rightarrow i}(p, c)|},
\end{equation}

SimpleRecon~\cite{sayed2022simplerecon} calculates cost volume using dot product of $F_i$. Accordingly, we define our feature-metric loss for \ours-M as the negative average dot product of $F_i$ and $F_{j\rightarrow i}$:
\begin{equation}
    \mathcal{L}_\text{feat}(i, j) = -\frac{1}{\vert I_{j\to i}\vert}\sum_{p\in I_{j\to i}} F_i(p) \cdot F_{j\rightarrow i}(p)
\end{equation}

The total loss $\mathcal{L}(i, j)$ is a sum of losses $\mathcal{L}(i, j) = \mathcal{L}_\text{photo}(i, j) + \mathcal{L}_\text{depth}(i, j) + \mathcal{L}_\text{feat}(i, j)$.

\section{EXPERIMENTS}
\label{sec:experiments}

\subsection{Datasets}

We evaluated our method on TUM RGB-D~\cite{sturm2012tum}, 7Scenes~\cite{shotton2013scenes}, and ScanNet~\cite{dai2017scannet} indoor benchmarks, following the evaluation protocol of DeepVideoMVS~\cite{duzceker2021deepvideomvs}. 
Moreover, we included ARKitScenes~\cite{dehghan2021arkitscenes} as an in-the-wild benchmark. This large-scale dataset, collected using a tablet with an online ARKit tracking system, features noisy camera poses and fast camera movements. We evaluate on the first 25 sequences longer than 20 seconds from the validation subset, for which the corresponding RGB and ground truth depth are available.

\begin{figure*}[ht!]
\centerline{
\begin{tabular}{cccccc}
Ground truth & CVD~\cite{luo2020cvd}  & RobustCVD~\cite{kopf2021rcvd} & GCVD~\cite{lee2022gcvd} & \ours-S (ours) & \ours-M (ours) \\
\includegraphics[width=0.145\linewidth]{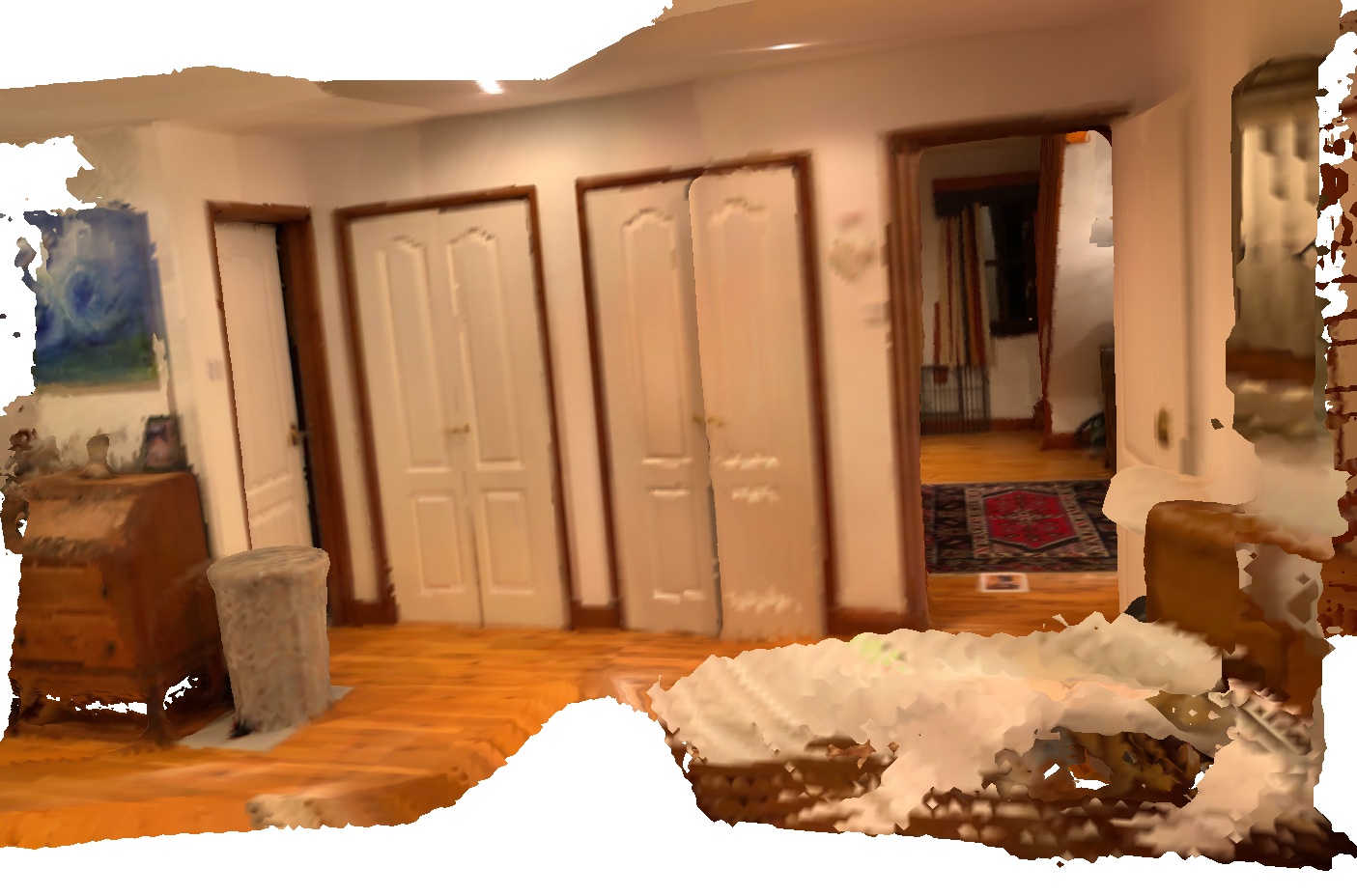} & 
\includegraphics[width=0.145\linewidth]{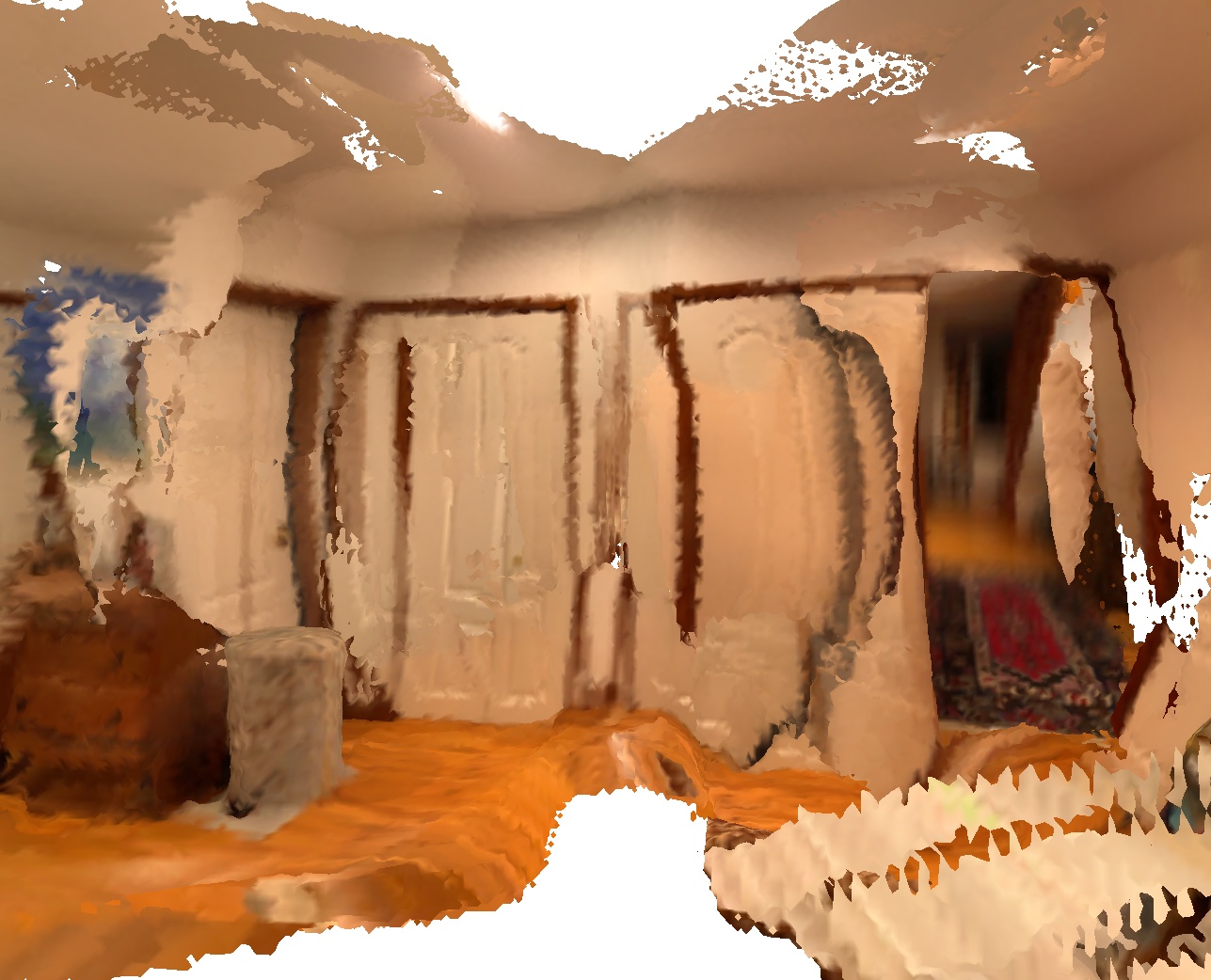} & 
\includegraphics[width=0.145\linewidth]{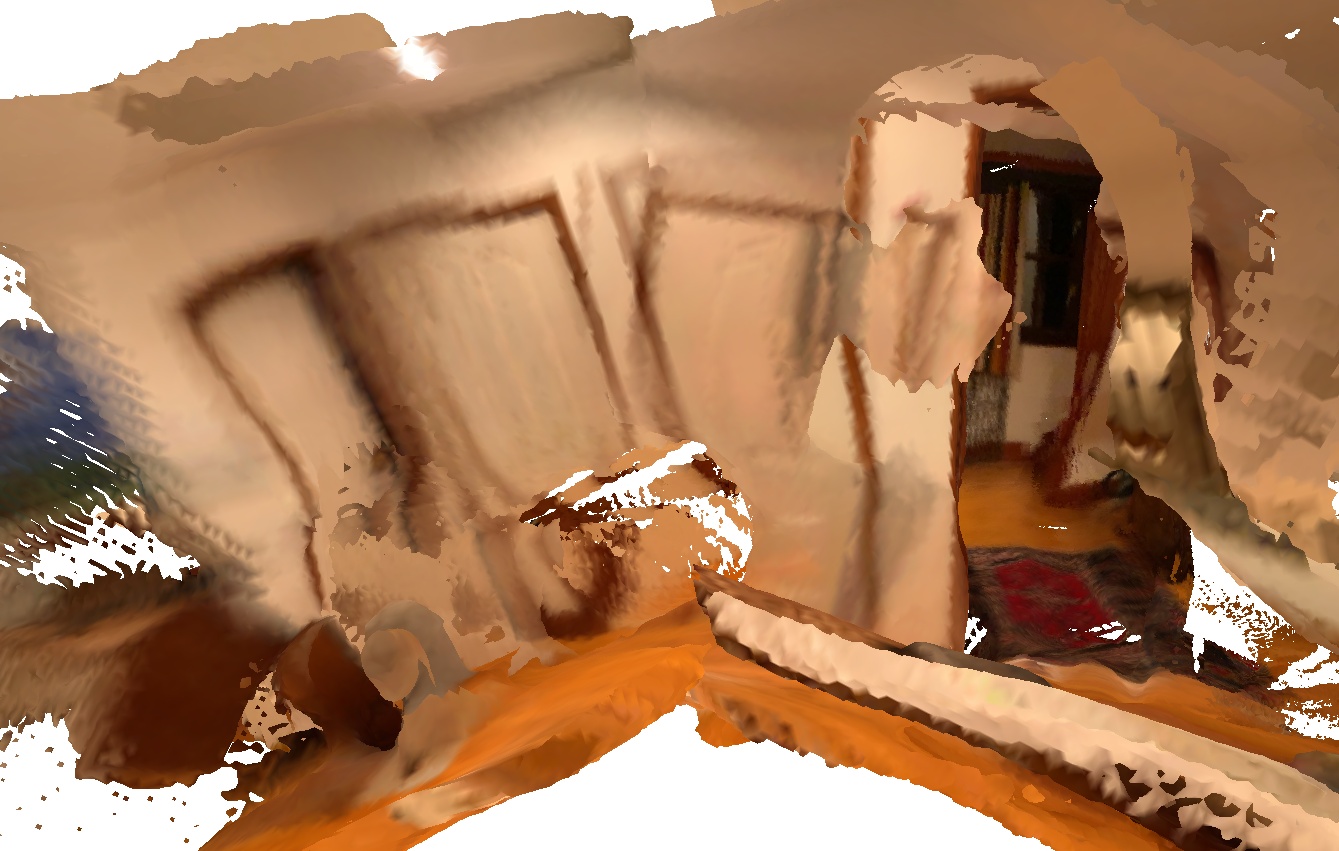} & 
\includegraphics[width=0.145\linewidth]{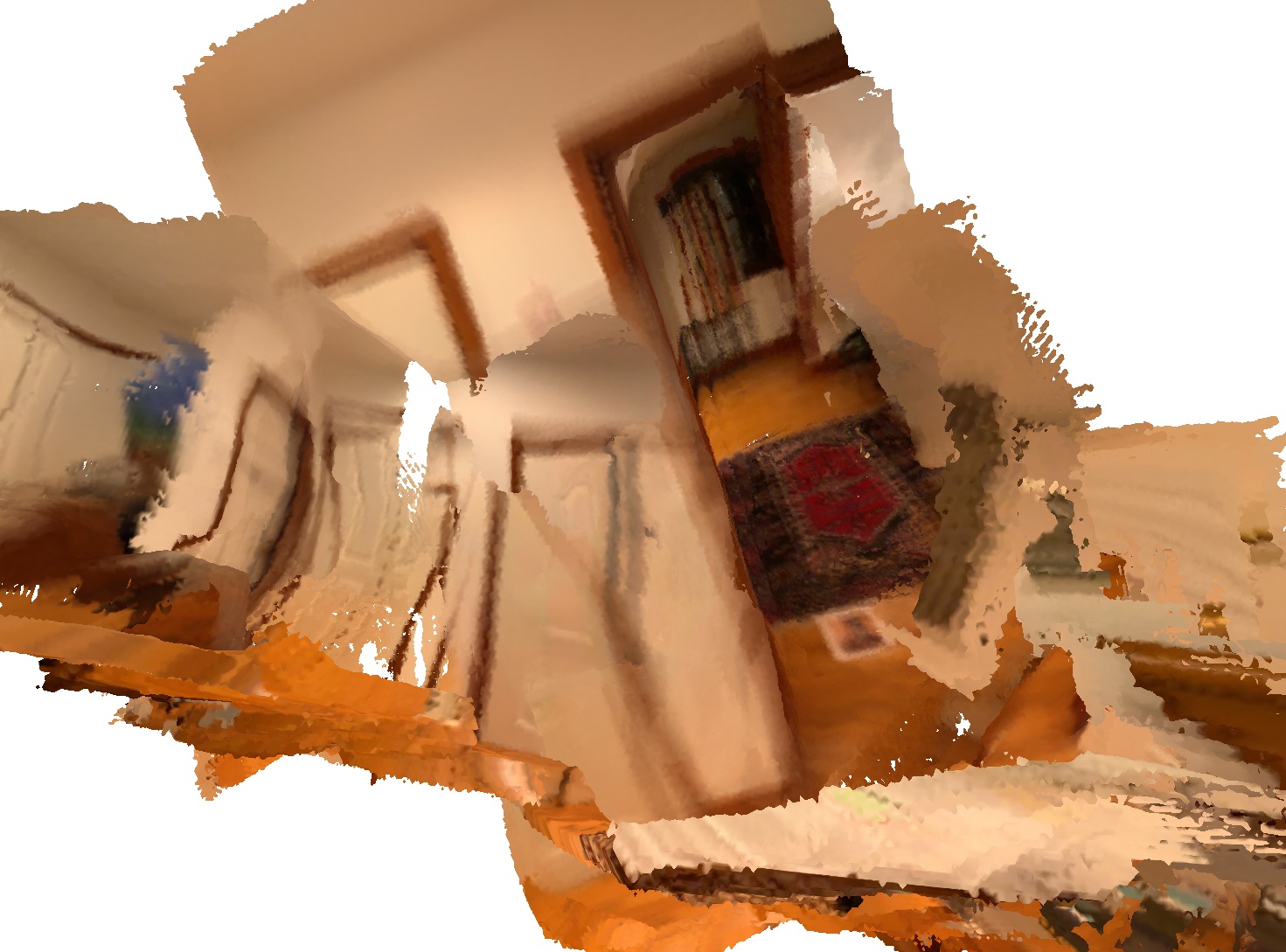} & 
\includegraphics[width=0.145\linewidth]{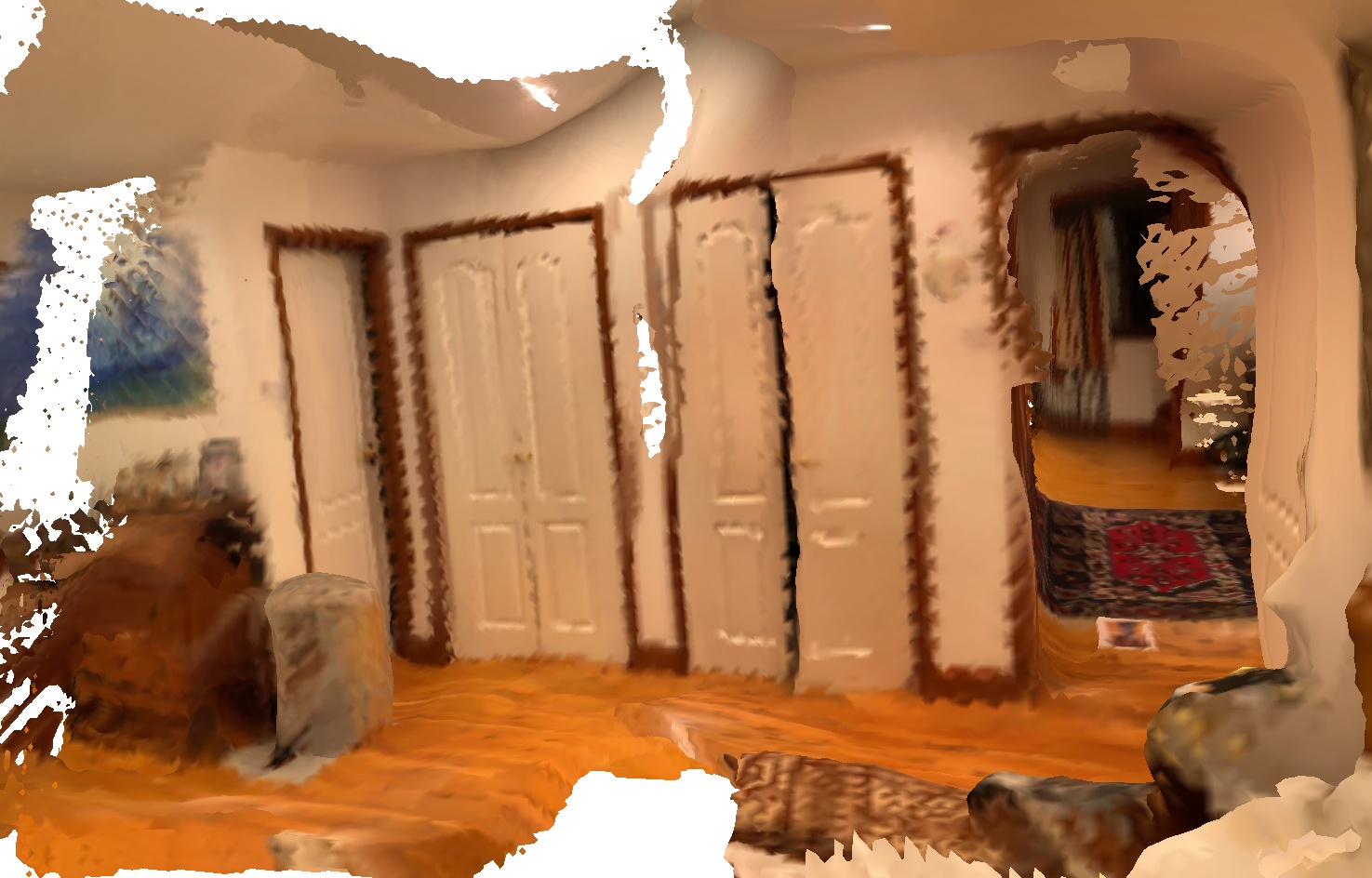} & 
\includegraphics[width=0.145\linewidth]{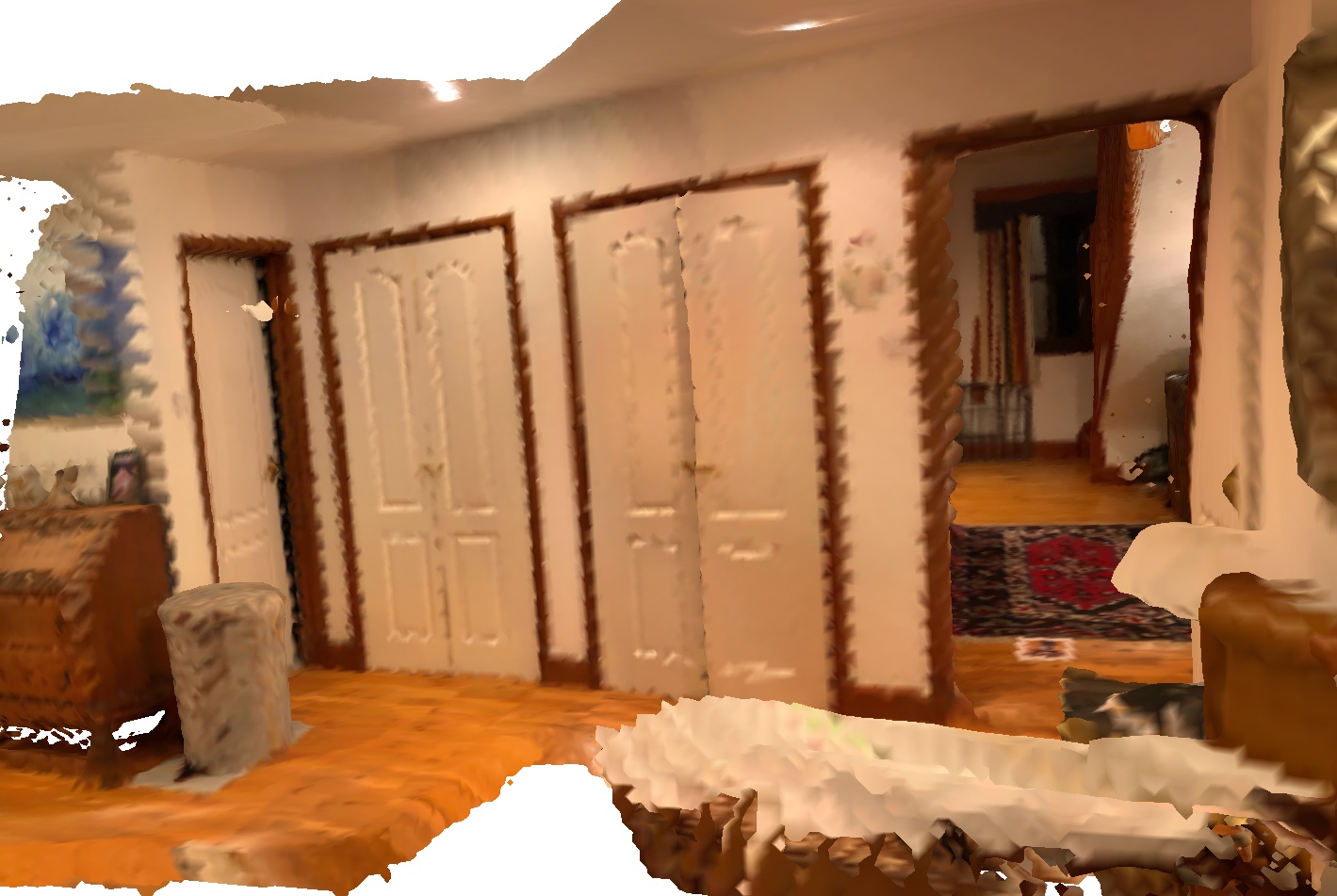} \\
\includegraphics[width=0.145\linewidth]{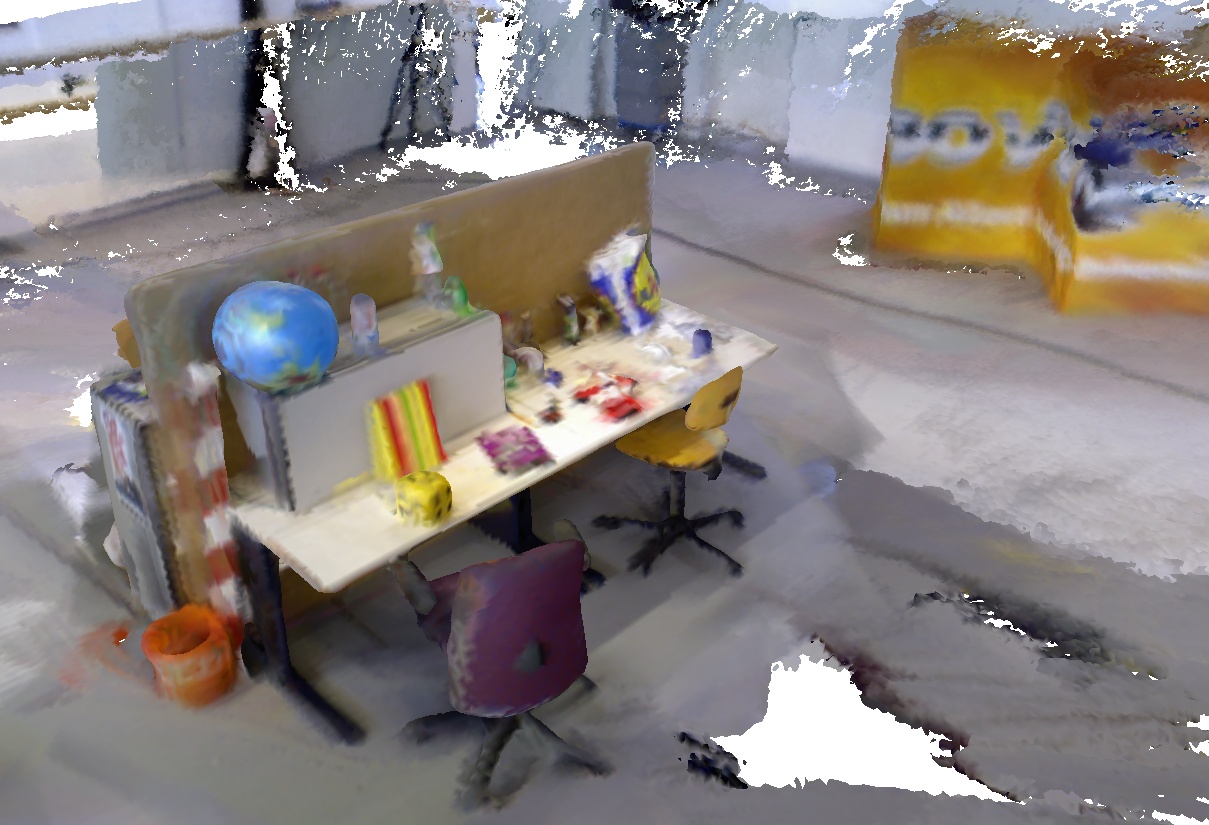} & 
\includegraphics[width=0.145\linewidth]{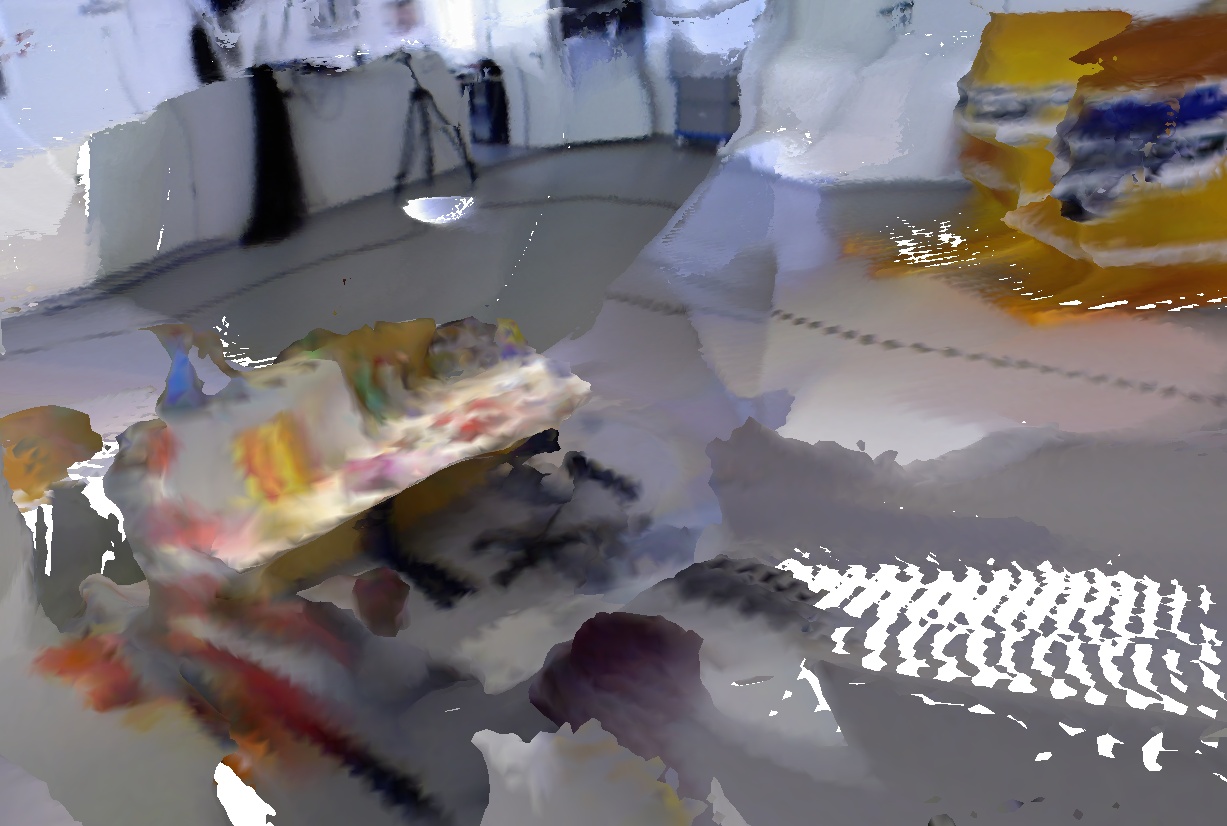} & 
\includegraphics[width=0.145\linewidth]{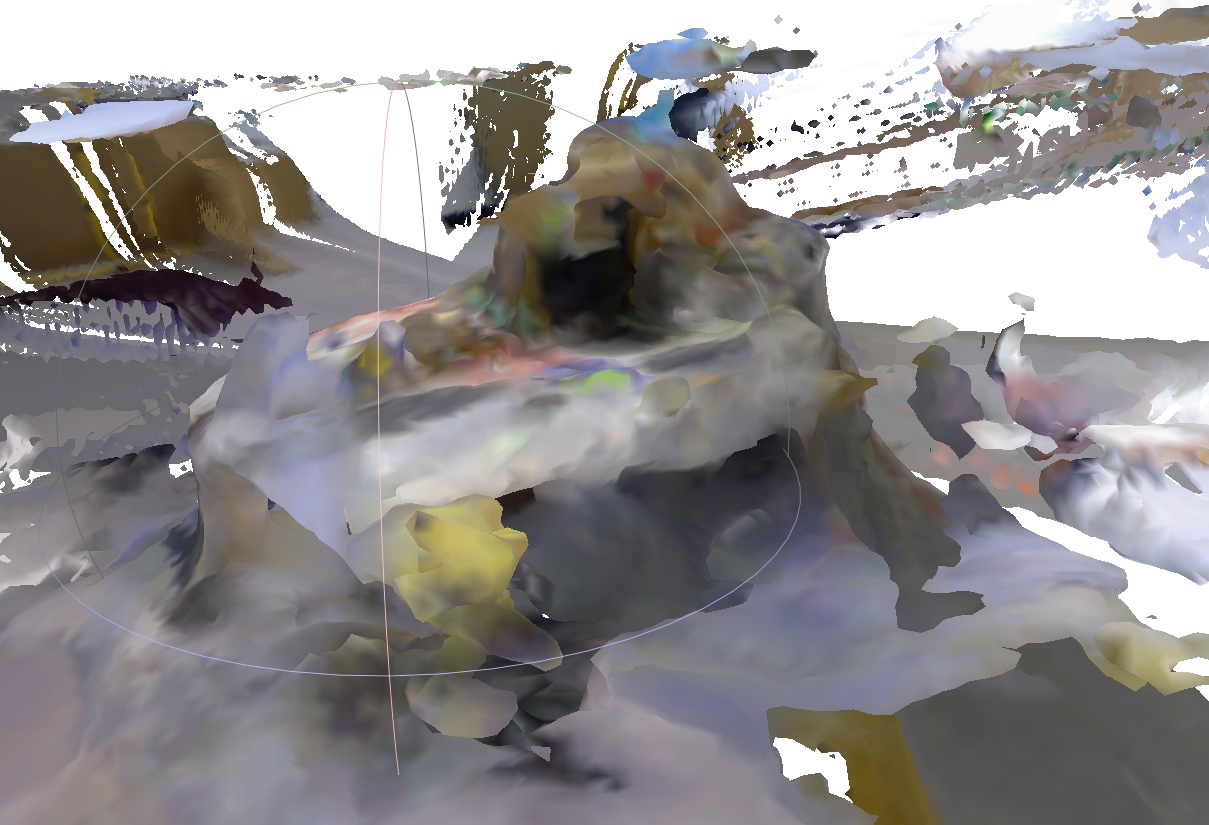} & 
\includegraphics[width=0.145\linewidth]{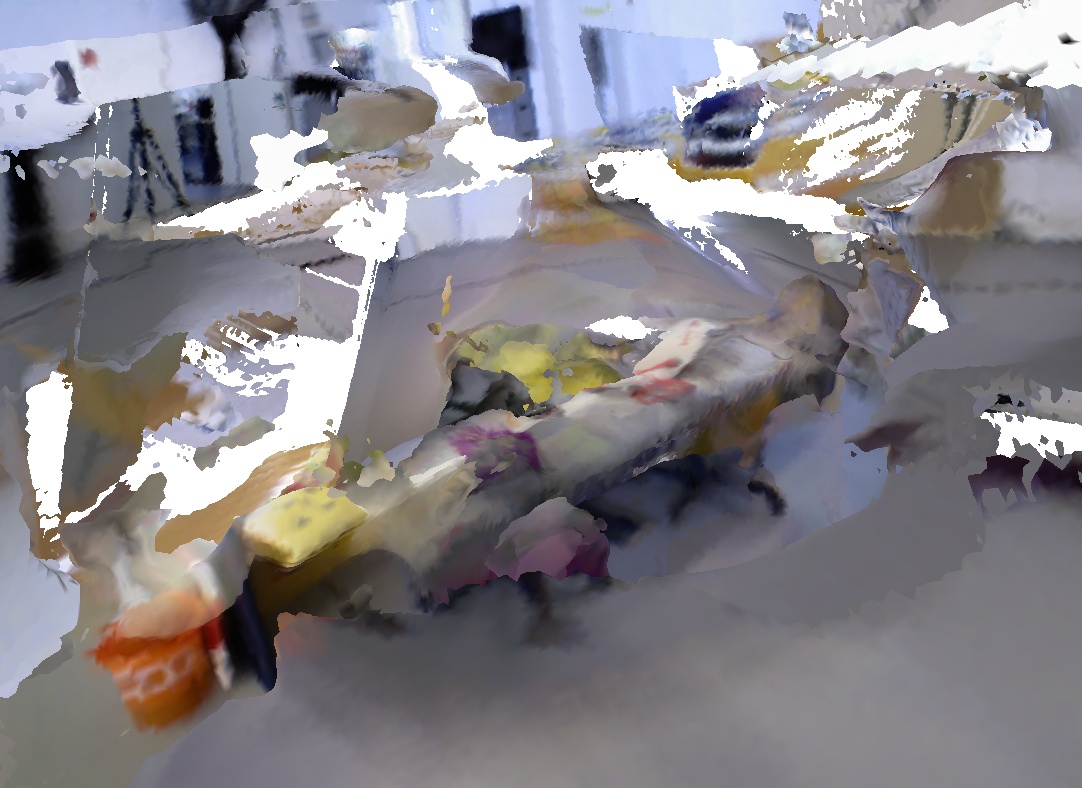} & 
\includegraphics[width=0.145\linewidth]{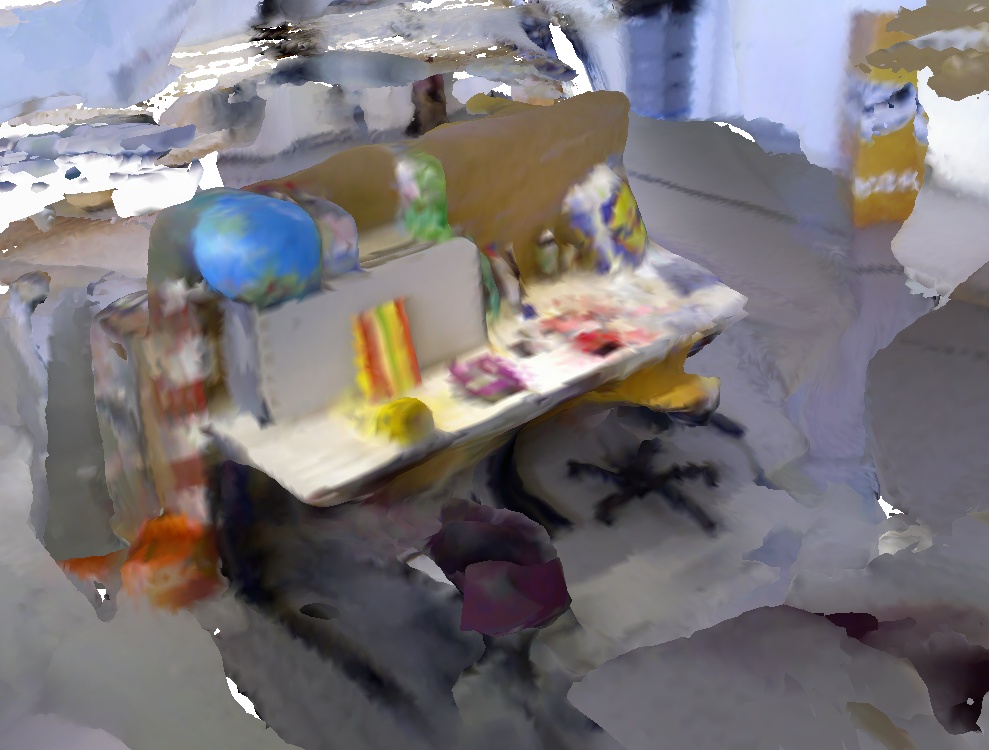} & 
\includegraphics[width=0.145\linewidth]{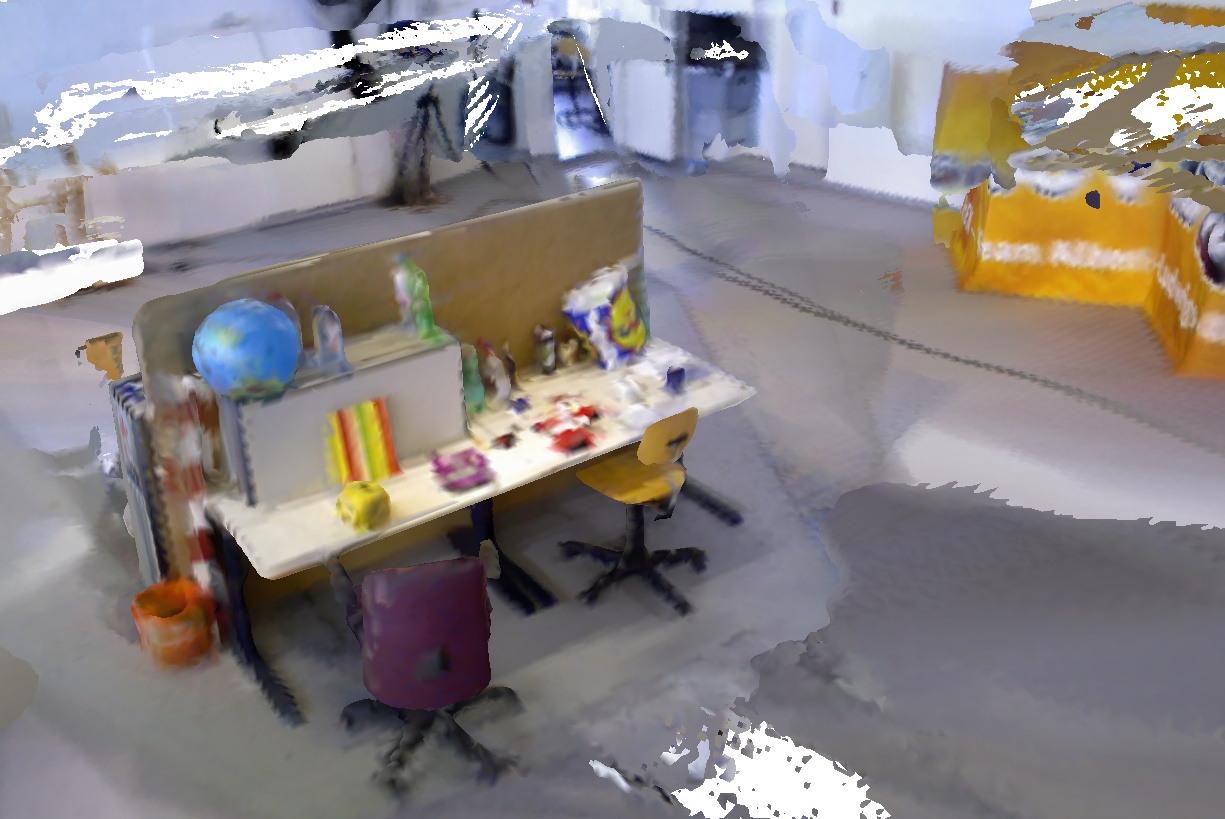} \\
\end{tabular}
}
\caption{ARKitScenes (top) and TUM RGB-D (bottom) scenes, reconstructed using depth maps produced by different test-time optimization methods, including \ours. For a fair comparison, we do not use ground truth camera poses but estimate them with DROID-SLAM~\cite{teed2021droidslam}. Obviously, other methods struggle to restore a general scene structure, while \ours{} provides well-aligned scans with fewer visual artifacts.}
\label{fig:tsdf}
\end{figure*}

\begin{table*}[ht!]
\caption{Results of metric depth estimation methods. No additional scale alignment is applied prior to this evaluation. The best scores are \textbf{bold}, the second best are \underline{underlined}.
}
\label{tab:results-metric}
\setlength{\tabcolsep}{4pt}
\centering
\begin{tabular}{clcccccccccccc}
\toprule
& \multirow{4}{*}{Method} & \multicolumn{3}{c}{TUM RGB-D} & \multicolumn{3}{c}{7Scenes} & \multicolumn{3}{c}{ScanNet} & \multicolumn{3}{c}{ARKitScenes}\\
\cmidrule{3-5} \cmidrule{6-8} \cmidrule{9-11} \cmidrule{12-14}
& & abs & abs & \multirow{3}{*}{\shortstack{$\delta_{1.25}$\\$[\%]\uparrow$}} & abs & abs & \multirow{3}{*}{\shortstack{$\delta_{1.25}$\\$[\%]\uparrow$}} & abs & abs & \multirow{3}{*}{\shortstack{$\delta_{1.25}$\\$[\%]\uparrow$}} & abs & abs & \multirow{3}{*}{\shortstack{$\delta_{1.25}$\\$[\%]\uparrow$}}\\
& & diff & rel & & diff & rel & & diff & rel & & diff & rel & \\
& & [m]$\downarrow$ & [\%]$\downarrow$ & & [m]$\downarrow$ & [\%]$\downarrow$ & & [m]$\downarrow$ & [\%]$\downarrow$ &  & [m]$\downarrow$ & [\%]$\downarrow$ &\\
\midrule
Single-view & ZoeDepth~\cite{bhat2023zoedepth}  & 0.461 & 20.6 & 65.7 & 0.349 & 23.2 & 66.5  & 0.600 & 53.0 & 31.7 & 0.286 & 19.5 & 75.0 \\
\midrule
\multirow{6}{*}{Multi-view}
& MVDepthNet~\cite{wang2018mvdepthnet}          & 0.290 & 11.7 & 86.0 & 0.203 & 11.6 & 86.9  & 0.165 & 8.5 & 92.8 & - & - & - \\
& DPSNet~\cite{im2019dpsnet}                    & 0.326 & 13.4 & 83.1 & 0.249 & 14.8 & 82.6  & 0.155 & 8.0 & 93.3 & - & - & - \\
& DELTAS~\cite{sinha2020deltas}                 & 0.353 & 12.7 & 81.8 & 0.191 & 11.4 & 88.2  & 0.150 & 7.9 & 93.8 & - & - & - \\
& GP-MVS~\cite{hou2019gpmvs}                    & \underline{0.244} & 10.4 & 88.9 & 0.196 & 11.8 & 87.2  & 0.149 & 7.6 & 94.0 & - & - & - \\
& DeepVideoMVS~\cite{duzceker2021deepvideomvs}  & 0.288 & 9.8  & 88.5 & 0.145 &  8.3 & 93.8  & 0.119 & 5.8 & 96.7 & - & - & - \\
& SimpleRecon~\cite{sayed2022simplerecon}       & 0.262 & \underline{7.9}  & \underline{89.6} & \underline{0.105} &  \underline{5.8} & \textbf{97.4}  & \textbf{0.087} & \underline{4.3} & \underline{98.1} & \underline{0.117} & \underline{7.8}& \underline{93.2} \\
\midrule
\multirow{2}{*}{\shortstack{Test-time\\optim.}}
& \ours-S (ours)                            & 0.515 & 17.0 & 73.0 & 0.158 & 8.3 & 91.0 & 0.193 & 9.5 & 89.5 & 0.182 & 12.2 & 84.5 \\
& \textbf{\ours-M (ours)}                              & \textbf{0.208} & \textbf{6.4}  & \textbf{93.0} &	\textbf{0.094} & \textbf{5.0}  & \underline{97.0} & \underline{0.088}	& \textbf{4.2} & \textbf{98.2} & \textbf{0.108} & \textbf{7.1} & \textbf{93.6} \\
\bottomrule
\end{tabular}
\end{table*}

\subsection{Evaluation Protocol}

We evaluate our approach in two set-ups. First, we focus on depth estimation solely, assuming that ground truth camera poses are available. Alternatively, we emulate usage without given camera poses and estimate up-to-scale camera poses with DROID-SLAM~\cite{teed2021droidslam}. The quality is measured with the standard metrics~\cite{ranftl2020midas,romanov2021gp,bhat2023zoedepth}: \textit{abs rel}, \textit{abs diff} and $\delta < 1.25$.

\subsection{Implementation Details}
All experiments are performed on a single Tesla P40 GPU.

In \ours-S, we extract features $F$ from the encoder block 3 of the backbone (chosen empirically). In \ours-M, we use the output of SimpleRecon~\cite{sayed2022simplerecon} feature encoder, as these features were trained to match across different frames.

Images are resized with a scale of 1/4. The size of log-scale tensors $l_i$ is $(8, 10)$ for horizontal and $(10, 8)$ for vertical videos. Before adjustment, $l_i$ are initialized with $\log(1) = 0$. Then, they are updated using Adam optimizer with a batch size of 128 frame pairs. The learning rate is set to 0.1 and degrades exponentially with $\gamma=0.996$ at Stage I and $\gamma=0.96$ at Stage II. Stage I and Stage II last for 600 and 60 epochs. The optimization stops when the loss does not reduce by at least 1\% throughout 40 epochs at the Stage I and 4 epochs at the Stage II.

\section{RESULTS}
\label{sec:results}

\begin{table*}[ht!]
\caption{Results of up-to-scale depth estimation methods, that do not have access to ground truth poses. \ours{} relies on camera poses from DROID-SLAM~\cite{teed2021droidslam}. Predicted depth maps are scale-aligned with ground truth ones prior to the evaluation.}
\label{tab:results-uts}
\centering
\begin{tabular}{clccccccccc}
\toprule
& \multirow{4}{*}{Method} & \multicolumn{3}{c}{TUM RGB-D} & \multicolumn{3}{c}{7Scenes} & \multicolumn{3}{c}{ARKitScenes}\\
\cmidrule{3-5} \cmidrule{6-8} \cmidrule{9-11}
& & abs diff & abs rel & $\delta_{1.25}$ & abs diff & abs rel & $\delta_{1.25}$ & abs diff & abs rel & $\delta_{1.25}$ \\
& & [m]$\downarrow$ & [\%]$\downarrow$ & [\%]$\uparrow$ & [m]$\downarrow$ & [\%]$\downarrow$ & [\%]$\uparrow$ & [m]$\downarrow$ & [\%]$\downarrow$ & [\%]$\uparrow$ \\
\midrule
Single-view & Romanov et al.~\cite{romanov2021gp} & 0.320 & 14.3 & 82.2 & 0.216 & 12.3 & 83.7 & 0.230 & 15.7 & 78.6 \\
\midrule
 \multirow{4}{*}{\shortstack{Test-time\\optim.}}
 & RobustCVD~\cite{kopf2021rcvd}                 & 0.452 & 19.3 & 66.8 & 0.403 & 21.7 & 56.8 & 0.387 & 23.8 & 63.9 \\
 & GCVD~\cite{lee2022gcvd}                       & 0.643 & 29.6 & 51.0 & 0.391 & 21.1 & 57.0 & 0.587 & 45.3 & 41.8 \\
 & \ours-S (ours)                             & \underline{0.221} & \underline{9.2} & \underline{89.2} &	\underline{0.137} & \underline{7.4} & \underline{92.5} & \underline{0.150} & \underline{9.9} & \underline{88.3} \\
 & \textbf{\ours-M (ours)}                             & \textbf{0.139}& \textbf{5.5} & \textbf{95.5} & \textbf{0.066}&	\textbf{3.5} & \textbf{97.6}& \textbf{0.096} & \textbf{6.1} & \textbf{93.7}\\
\bottomrule
\end{tabular}
\end{table*}

\subsection{Comparison to Prior Work}

\subsubsection{Competitors}

We evaluate \ours{} against various depth estimation approaches. ZoeDepth~\cite{bhat2023zoedepth} (metric) and Romanov et al.~\cite{romanov2021gp} (up-to-scale) represent the single-view paradigm. Then, we report results of MVS approaches~\cite{duzceker2021deepvideomvs, im2019dpsnet, wang2018mvdepthnet,sinha2020deltas,hou2019gpmvs}, including SimpleRecon~\cite{sayed2022simplerecon}.

\begin{table}[ht!]
\caption{Comparison with CVD~\cite{luo2020cvd} on TUM RGB-D sequences where COLMAP succeeded. \ours{} uses camera poses from DROID-SLAM~\cite{teed2021droidslam}. Predicted depth maps are scale-aligned with ground truth ones prior to the evaluation.}
\label{tab:results-cvd}
\setlength{\tabcolsep}{3pt}
\centering
\begin{tabular}{lccccc}
\toprule
\multirow{2}{*}{Method} & abs diff & abs rel & $\delta_{1.25}$ & Time per \\
& [m]$\downarrow$ & [\%]$\downarrow$ & [\%]$\uparrow$ & frame [sec]$\downarrow$ \\
\midrule
CVD~\cite{luo2020cvd}           & \underline{0.200} & \underline{8.7} & \underline{90.5} & 74.9 \\
RobustCVD~\cite{kopf2021rcvd}   & 0.312 & 14.1 & 82.5 & 31.6 \\
GCVD~\cite{lee2022gcvd}         & 0.315 & 14.7 & 80.4 & 10.0 \\
\ours-S (ours)                  & 0.215 & 8.9 & 89.5 & \textbf{0.44} \\
\textbf{\ours-M (ours)}         & \textbf{0.128} &	\textbf{5.6} & \textbf{94.9} & \underline{0.88} \\
\bottomrule
\end{tabular}
\end{table}

We also compare against test-time optimization approaches. CVD~\cite{luo2020cvd} uses COLMAP poses and predicts up-to-scale depth. RobustCVD~\cite{kopf2021rcvd} and GCVD~\cite{lee2022gcvd} estimate camera poses and depth maps jointly, hence providing up-to-scale predictions. Surprisingly, these methods do not benefit from using ground truth poses: we attribute this to the joint training procedure, causing depth estimation errors to be partially compensated with camera pose drifts.

\subsubsection{Quantitative Results}

Tab.~\ref{tab:results-metric} proves test-time optimization to improve over state-of-the-art MVS approaches. Evidently, MVS strategies of aggregating spatial information are suboptimal, and there is still room for enhancement. Since consistency is a prerequisite for accuracy, addressing inconsistency with a direct optimization improves individual depth maps. The gain w.r.t the single-view metric ZoeDepth~\cite{bhat2023zoedepth} method is even more tangible, especially on 7Scenes and ScanNet. SimpleRecon~\cite{sayed2022simplerecon} is trained on ScanNet, so the gain is more prominent on other datasets, proving better generalization of \ours{}.

In Tab.~\ref{tab:results-uts}, we compare test-time optimization approaches. As they predict depth up to scale, we align the depth maps with ground truth ones before computing metrics. We omit ScanNet here, since the test-time optimization would take up to six GPU-months for competing approaches~\cite{luo2020cvd, kopf2021rcvd, lee2022gcvd}. Evidently, \ours{} outperforms others by a large margin, and sets a new strong state-of-the-art in video depth estimation. Besides, \ours{}-S demonstrates a solid gain over initial single-view predictions by Romanov et al.~\cite{romanov2021gp}, proving the viability of our approach. 

We also evaluate against the seminal test-time optimization CVD~\cite{luo2020cvd} on TUM RGB-D. CVD~\cite{luo2020cvd} estimates camera poses with COLMAP, which succeeds on 8 out of 13 test videos~\cite{duzceker2021deepvideomvs}, so we compare other test-time optimization methods on the same videos. As shown in Tab.~\ref{tab:results-cvd}, apart from superior depth estimation quality, \ours{} is significantly faster than the competing approaches.

\subsubsection{Qualitative Results}

Consistent depth maps allow obtaining precise 3D scans, while any severe inconsistencies would reveal themselves clearly as reconstruction artifacts. To demonstrate the superior consistency of depth maps produced by \ours{} in comparison with other consistent depth estimation methods, we use these maps to reconstruct scenes via TSDF fusion, ans visualize the scans in Fig.~\ref{fig:tsdf}. As one can see from a side-by-side comparison, our depth adjustment approach provides well-aligned reconstructions with a clear scene structure. 

\subsection{Ablation Study}

To provide a comprehensive study of \ours{}, we run additional experiments focusing on different aspects. In the ablation study, we use TUM RGB-D, since this benchmark is arguably the most commonly used and hence well-studied.

\begin{table}[ht!]
\caption{Ablation study of losses on TUM RGB-D.}
\label{tab:ablation-losses}
\centering
\begin{tabular}{lccccccccc}
\toprule
\multirow{2}{*}{Method} & abs diff & abs rel & $\delta_{1.25}$ & Time per\\
 & [m]$\downarrow$ & [\%]$\downarrow$ & [\%]$\uparrow$ & frame [sec] \\
\midrule
\ours-M & \textbf{0.208} &\textbf{6.4}&\textbf{93.0} & 0.88 \\
\quad w/o $\mathcal{L}_\text{photo}$ & \underline{0.213} & \underline{6.5} & \underline{92.8} & 0.86 \\
\quad w/o $\mathcal{L}_\text{depth}$ &0.386 & 15.2 & 77.5 & 0.85 \\
\quad w/o $\mathcal{L}_\text{feat}$ & 0.285 & 9.1 & 88.7 & 0.85 \\
\bottomrule
\end{tabular}
\end{table}

\begin{table}[ht!]
\caption{Effect of depth scale propagation on TUM RGB-D.}
\label{tab:ablation-optimization}
\centering
\begin{tabular}{lccc}
\toprule
\multirow{2}{*}{Method} & abs diff & abs rel & $\delta_{1.25}$ \\
 & [m]$\downarrow$ & [\%]$\downarrow$ & [\%]$\uparrow$ \\
\midrule
\ours-M & \textbf{0.208} & \textbf{6.4}&\textbf{93.0}  \\
\quad w/o scale propagation & 0.284& 9.1 & 88.7 \\
\bottomrule
\end{tabular}
\end{table}

\subsubsection{Losses}

We run \ours{} with different loss terms. According to the Tab.~\ref{tab:ablation-losses}, both depth and feature-metric losses have a major impact on the accuracy. The photometric loss does not add much to the performance: it might be redundant when feature-metric loss is applied, as features already contain sufficient information about objects' appearance. Using all losses ensures the best results, while the computational overhead is minor.

\subsubsection{Depth Scale Propagation}

To prove our depth scale propagation, we try initializing the non-keyframe depth scale maps with ones, same as for keyframes at the Stage I. Evidently, turning our novel depth scale propagation scheme off worsens all the scores, while the improvement of inference speed is negligible (Tab.~\ref{tab:ablation-optimization}). So, we can confidently conclude that depth estimation benefits from initializing depth scale maps by reprojection.

\section{CONCLUSION}
\label{sec:conclusion}

We presented \ours{}, that addresses depth estimation from videos. \ours{} obtains depth priors with a pre-trained network, and performs inference-time optimization to ensure depth consistency between frames. By using image features to guide the optimization and applying the novel scale propagation strategy, our approach generates high-quality depth maps an order of magnitude faster than the previous state-of-the-art test-time optimization approach. \ours{} outperformed existing single-view, multi-view, and test-time depth estimation approaches on the standard benchmarks and also proved to handle imperfect smartphone-captured data. Overall, we set a new state-of-the-art in video depth estimation in both accuracy and efficacy.



\vfill\pagebreak



\bibliographystyle{IEEE_ICIP2024}
\bibliography{references}

\end{document}